%% file: main.tex
\documentclass[11pt,a4paper]{article}
\PassOptionsToPackage{breaklinks}{hyperref}
\usepackage[hyperref]{naaclhlt2019}
\usepackage{times}
\usepackage{latexsym}
\usepackage{booktabs}
\usepackage{multirow}
\usepackage{url}
\usepackage{amsmath}
\usepackage{amsfonts}
\usepackage{color,soul}
\usepackage{mdframed}
\usepackage{xcolor}
\usepackage[color=gray,disable]{todonotes}

\colorlet{Changes@Color}{red}

\aclfinalcopy 
\def\aclpaperid{2387} 

\title{Generating Token-Level Explanations for Natural Language Inference} 

\author{James Thorne \\
  University of Cambridge \\
  {\tt jt719@cam.ac.uk} \\\And
  Andreas Vlachos \\
  University of Cambridge \\
  {\tt av308@cam.ac.uk} \\\AND
  Christos Christodoulopoulos \\
  Amazon \\
  {\tt chrchrs@amazon.co.uk} \\\And
  Arpit Mittal \\
  Amazon \\
  {\tt mitarpit@amazon.co.uk} \\
  }

\date{}

\begin{document}
\maketitle
\begin{abstract}
\todo{Untriaged}
\todo[color=red]{Show stopper}
\todo[color=yellow]{Pending}
\todo[color=green]{Maybe solved: please review}
\todo[color=cyan]{Solved}
\todo[color=white]{Redundant}
The task of Natural Language Inference (NLI) is widely modeled as supervised sentence pair classification. While there has been a lot of work recently on generating explanations of the predictions of classifiers on a single piece of text, there have been no attempts to generate explanations of classifiers operating on pairs of sentences. In this paper, we show that it is possible to generate token-level explanations for NLI without the need for training data explicitly annotated for this purpose. We use a simple LSTM architecture and evaluate both LIME and Anchor explanations for this task. We compare these to a Multiple Instance Learning (MIL) method that uses thresholded attention make token-level predictions. The approach we present in this paper is a novel extension of zero-shot single-sentence tagging to sentence pairs for NLI. We conduct our experiments on the well-studied SNLI dataset that was recently augmented with manually annotation of the tokens that explain the entailment relation. We find that our white-box MIL-based method, while orders of magnitude faster, does not reach the same accuracy as the black-box methods.
\end{abstract}

\input{1_intro}
\input{3_model}

\input{4_experiments}

\input{5_discussion}

\section*{Acknowledgements}
This work was conducted while James Thorne was an intern at Amazon. Andreas Vlachos is supported by the EU H2020 SUMMA project (grant agreement number 688139).

\bibliography{refs}
\bibliographystyle{acl_natbib}

\end{document}

%% file: 1_intro.tex
\section{Introduction}
Large-scale datasets for Natural Language Inference (NLI) \cite{Bowman2015,Williams2018} have enabled the development of many 
deep-learning 
models 
\citep{Rocktaschel2016,Peters2018,Radford2018}. 
The task is modeled as 3-way classification of the entailment relation between a pair of sentences. Model performance is assessed through 
accuracy 
on a held-out 
test set. 
While state-of-the-art models achieve high accuracy, their complexity makes it difficult to interpret their behavior. 
\begin{figure}
\begin{tabular}{p{2cm}p{5cm}}
\textbf{Premise}:    & Children {\underline{\hl{smiling}}} and waving at a camera \\
\textbf{Hypothesis}: & The kids are {\underline{\hl{frowning}}}                   \\
\textbf{Label}:      & Contradiction               
\end{tabular}
\caption{Example of token-level highlights from the e-SNLI dataset \cite{Camburu2018}. Annotators were provided a premise and hypothesis 
and asked to highlight words considered essential to explain the label.} 
\label{fig:example}
\end{figure}

Explaining the predictions made by classifiers has been of increasing concern \citep{Doshi-Velez2017}.
It has been studied in natural language processing through both black-box analysis, and through modifications to the models under investigation; we refer to the latter approaches as \emph{white-box}. 
Common black-box techniques generate 
explanations of predictions through training meta-models by perturbing input tokens \citep{Ribeiro2016,Nguyen2018,Ribeiro2018} or through interpretation of model sensitivity to input tokens \citep{Li2016, Feng2018}. 
White-box methods induce new features \citep{Aubakirova2016}, augment models to generate explanations accompanying their predictions \citep{Lei2016, Camburu2018}, or expose model internals such as magnitude of hidden states \citep{Linzen2016}, gradients (as a proxy for model sensitivity to input tokens \citep{Li2016}) or attention \citep{Bahdanau2014,Xu2015}.

Model explanations typically comprise a list of features (such as tokens) that contributed to the prediction and can serve two distinct purposes: acting either as a diagnostic during model development or to allow for a rationale to be generated for a system user.
While methods for explaining predictions may output what was salient to the model, there is no guarantee these will correspond to the features that users deem important.

In this paper we introduce a white-box method that thresholds the attention matrix of a neural entailment model to induce token-level explanations. To encourage the model's prediction of salient tokens to correspond better to the tokens users would find important, our approach uses Multiple Instance Learning (MIL) \citep{maron1998framework} to regularize the attention distributions. 

We compare this against two black-box methods: LIME \citep{Ribeiro2016} and Anchor Explanations \citep{Ribeiro2018}; both white- and black-box methods are applied to a 
simple neural architecture relying on independent sentence encoding with cross-sentence attention, and thus could also be applied to more complex architectures of the same family. 
Finally, we also compare against a fully supervised baseline trained to jointly predict entailment relation and token-level explanations. Our experiments are conducted on e-SNLI \cite{Camburu2018}, a recently introduced extension to SNLI \citep{Bowman2015}, containing human-selected highlights of which words are required to explain the entailment relation between two sentences (see Fig.~\ref{fig:example}).

Our experimental results indicate that regularizing the model's attention distributions encourages the explanations generated to be better aligned with human judgments (even without our model having explicit access to the labels which tokens annotators found important). Compared to the baseline thresholded attention mechanism, our method provides an absolute increase in token-level precision and recall by $6.68\%$ and $28.05\%$ respectively for the hypothesis sentence for e-SNLI explanations. 

We also found that attention based explanations are not as reliable as black-box model explanation techniques, as indicated by higher $F_1$ scores for both LIME and Anchor Explanations. This is consistent with findings of contemporaneous work by \citet{jain2019attention}. However, we do show that, if generating explanations from a model is a requirement, incorporating an explicit objective in training can be beneficial. 
This can be particularly useful in practicw 
due to the computational cost of black-box model explanations, which in empirical evaluation we found to be orders of magnitude slower ($0.01$ seconds vs $64$ seconds per instance). 

%% file: 3_model.tex
\section{NLI Model}
\label{sec:model}
The model we use for both white- and black-box experiments is based on an architecture widely adopted for sentence-pair classification \citep{Lan2018}. 
It comprises the following: 




\paragraph{Word Embeddings} We use pretrained GloVe embeddings \cite{Pennington2014} that were fixed during training.

\paragraph{Sentence Encoding} Both the 
premise and hypothesis are independently encoded with the same LSTM \cite{Hochreiter1997}, yielding $\mathbf{h}^p$ and $\mathbf{h}^h$ respectively.

\paragraph{Attention} A matrix of soft alignments between tokens in the premise sentence and the hypothesis sentence is computed using 
attention \cite{Bahdanau2014} over the encodings. 
Like \citet{Parikh2016}, we project the encoded sentence representations using a feed-forward network, $f_{attend}$,  ($u_i = f_{attend}(h^{p}_i)$, $v_j = f_{attend}(h^{h}_j)$) before computing the inner product: $\tilde{A}_{ij} = u_i^T v_j$. Given a premise of length $m$, the attention distribution for the hypothesis sentence is $\mathbf{a}^{h} = \text{normalize}(\tilde{A}_{m,*})$ where linear normalization is applied ($\text{normalize}(w)=\frac{w}{\|w\|_1}$). Likewise for the corresponding hypothesis of length $n$, the premise attention distribution is $\mathbf{a}^{p} = \text{normalize}(\tilde{A}_{*,n})$. 


\paragraph{Output Classifier} We predict the class label through a feed-forward neural network, $f_{cls}$, where both attended encodings of the premise 
and hypothesis final hidden states are concatenated as input: $f_{cls}( \lbrack a^{p}_mh^p_m ; a^{h}_nh^h_n \rbrack )$. The logits are normalized using the softmax function, yielding a distribution over class labels $\hat{y}$.  

\paragraph{Training} The model is trained in a supervised environment using cross-entropy loss between the predicted class labels for an instance $\hat{y}$ and the labeled value in the dataset, formally defined in Section~\ref{sec:gen}.

\section{Generating Token-Level Explanations}
\label{sec:gen}
Let $\mathbf{x}^p = (x^p_1, \ldots, x^p_m)$ and $\mathbf{x}^h = (x^h_1, \ldots, x^h_n)$ be sequences of tokens of length $m$ and $n$ respectively for the input premise and hypothesis sentences. Let $y$ represent an entailment relation between $\mathbf{x}^p$ and $\mathbf{x}^h$ where $y \in \{\text{entails}, \text{contradicts}, \text{neutral}\}$. 
Labeled training data is provided of the form $\{(\mathbf{x}^p_k, \mathbf{x}^h_k, y_k) \}^K_{k=1}$. For each instance, the model must generate an explanation $\mathbf{e}$ defined as a subset of zero or more tokens from both the premise and hypothesis sentences: $\mathbf{e}^p \in \mathcal{P}(\mathbf{x}^p)$, $\mathbf{e}^h \in \mathcal{P}(\mathbf{x}^h)$.

We generate token-level explanations by thresholding token attention weights. Concretely, we select all tokens, $x$, with a weight greater than a threshold. While similar to \citet{Rei2018}, we incorporate a re-scaling using the $\tanh$ function: 
$\mathbf{e}^p = \{x^p_i \vert \tilde{a}^p_i \in \tilde{A}_{*,n} \wedge \tanh(\tilde{a}^p_{i}) \ge \tau\}$ and likewise for the hypothesis.

\subsection{Multiple Instance Learning}
\label{sec:mil}
Thresholding the attention distributions from our model will give an indication of which tokens the model is weighting strongly for the entailment task. However, as mentioned in the introduction, there is no guarantee that this method of explaining model behavior will correspond with tokens that humans judge as a reasonable explanation of entailment. 
To better align the attention-based explanations with the human judgments, we model the generation of explanations as Multiple Instance Learning (MIL) \citep{maron1998framework}. In training the model sees labeled ``bags'' (sentences) of unlabeled features (tokens) and learns to predict labels both for the bags and the features. 
In MIL, this is often 
achieved by introducing regularizers when training. 
%
To encourage our NLI model to predict using sparser attention distributions (which we expect to correspond more closely with human token-level explanations), we introduce the following regularizers into the loss function:




\paragraph{$R_1$:} This entropy-based term allows us to penalize a model that uniformly distributes probability mass between tokens.
\begin{equation}
\begin{split}
R_1 & =  \sum^K_{k=1} \big( \mathbb{H}(\mathbf{a}^p_k) + \mathbb{H}(\mathbf{a}^h_k)\big)\\
    & = - \sum^K_{k=1} (\sum_{i=1}^m a^p_{k,i} \log a^p_{k,i} + \sum_{j=1}^n a^h_{k,j} \log a^h_{k,j})
\end{split}
\end{equation}
\paragraph{$R_2$:}
This term, adapted from a loss function for zero-shot tagging on single sentences \citep{Rei2018}, penalizes the model for breaking the assumption that at least one token must be selected from both premise and hypothesis sentences to form an explanation. 
The only exception is that, following the e-SNLI dataset annotation 
by \citet{Camburu2018}, if the neutral entailment is predicted, no tokens are selected from the premise. 
\begin{equation}
\begin{split}
R_2 = \sum^K_{k=1} \big(&(\max_i a^p_{k,i}-\mathbb{I}\lbrack y_k \neq \text{neutral} \rbrack)^2 \\ + &(\max_j a^h_{k,j} - 1)^2\big)
\end{split}
\end{equation}
\paragraph{$R_3$:} This term, also adapted from \citet{Rei2018}, encodes  
the assumption that not all tokens must be selected in the explanation. This is achieved by
penalizing the smallest non-zero attention weight, which has the effect of encouraging at least one weight to be close to zero.
\begin{equation}
R_3 = \sum^K_{k=1} \big( (\min_i a^p_{k,i})^2 + (\min_j a^h_{k,j})^2 \big)
\end{equation}
The loss function used for training of our proposed model
incorporating the regularizers which are controlled with hyperparameters is:
\begin{equation}
L = \sum_{k=1}^K{\sum_{c\in C}{y_{k,c} \log \hat{y}_{k,c}}} + \alpha R_1 + \beta R_2 + \gamma R_3
\end{equation}


\begin{table*}[th!]
\centering
\begin{tabular}{@{}l|c|ccc|ccc@{}}
\toprule
\multirow{3}{*}{\textbf{Model}} & \textbf{Runtime (s)}            & \multicolumn{6}{c}{\textbf{Token Explanation (\%)}}                                                                                                                                          \\
                                & \multirow{2}{*}{\textbf{per instance}} & \multicolumn{3}{c}{\textbf{Premise}}                                                    & \multicolumn{3}{c}{\textbf{Hypothesis}}                                                 \\
                                &                                    & \multicolumn{1}{c}{\textbf{P}}           & \multicolumn{1}{c}{\textbf{R}} & \multicolumn{1}{c}{\textbf{F1}} & \multicolumn{1}{|c}{\textbf{P}}           & \multicolumn{1}{c}{\textbf{R}} & \multicolumn{1}{c}{\textbf{F1}} \\ \toprule
Fully Supervised LSTM-CRF & 0.02 & 86.91 & 40.98 & 55.70 & 81.16 & 54.79 & 65.41 \\
\midrule\midrule
Thresholded Attention (Linear) & \textbf{0.01} & 19.96 & 19.67 & 19.56 & 46.70 & 34.92 & 39.89 \\
+ MIL Regularizers (R1) & - & 16.59 & 15.67 & 16.12 & 50.02 & 42.44  & 46.01 \\ 
+ MIL Regularizers (R2 + R3) & -  & 18.19 & 20.18 & 19.13 & 51.29 & 50.73 & 51.00  \\ 
+ MIL Regularizers (R1 + R2 + R3) & -  & 19.23 & 26.21 & 22.18 & 53.38 & 62.97 & 57.78 \\ 
\midrule
LIME & 64 & \textbf{60.56} & \textbf{48.28} & \textbf{53.72} & \textbf{57.04} & \textbf{66.92} & \textbf{61.58} \\
Anchors & 10 & 42.06 & 20.04 & 27.14 & 53.12 & 63.94 & 58.03 \\ 
\bottomrule
\end{tabular}
\caption{Token-level scores for human-selected explanations of NLI using the e-SNLI dataset. The select-all baseline precision for the premise is $18.5\%$ and $35.2\%$ for the hypothesis. \label{tab:res}}

\end{table*}

\section{Alternative Models}



\subsection{Black-box explanations of NLP models}
We use two established black-box model explanation techniques for generating token-level explanations: LIME \cite{Ribeiro2016} and Anchors \cite{Ribeiro2018}. Both 
techniques probe a classifier by making perturbations to a single input and modeling which of these perturbations influence the classification. To adapt these for use in NLI, we make a simple modification that runs the 
analysis twice: once for the premise sentence and once for the hypothesis sentence on the NLI model described in Section~\ref{sec:model}. 

\paragraph{LIME} Generates local explanations for a classifier through the introduction of a simple meta-model that is trained to replicate a local decision boundary of an instance under test. The training data is generated through observing the impact on classification when removing tokens from the input string. 

\paragraph{Anchor Explanations} Considers the distribution of perturbed instances in the neighborhood of the instance under test through word substitution to identify a rule (a set of tokens in our case) for which the classification remains unchanged.

\subsection{Supervised Model}
For a supervised model we build upon the model discussed in Section~\ref{sec:model}, adding components to support LSTM-CRF-based tagging \cite{Lample2016}. We use the following architecture:

\paragraph{Context Encoding} We use the same pretrained GloVe embeddings \citep{Pennington2014} that were fixed during training. The premise and hypothesis sentence were independently encoded with the same LSTM \citep{Hochreiter1997} yielding $\mathbf{h}^p$ and $\mathbf{h}^h$ respectively and attended to as per the description in Section~\ref{sec:model}.


\paragraph{Outputs} \todo{updated} The model is jointly trained with two output objectives: a labeling objective and a tagging objective. 
During training, the losses for both tasks are equally weighted. The first output objective is the three-way SNLI classification over the pair of sentences. This is the same component as the model presented in Section~\ref{sec:model}.

The second objective is a binary tagging objective over the highlighted token-level explanations. We use a jointly-trained LSTM-CRF decoder architecture \cite{Lample2016} which operates a CRF over encoded representations for each token. In our model, we independently decode the premise and hypothesis sentences. The inputs to our CRF are the attended premise and hypothesis: $\mathbf{a}^p \odot \mathbf{h}^p$ and $\mathbf{a}^h \odot \mathbf{h}^h$ respectively (where $\odot$ is the point-wise multiplication between the attention vector and the encoded tokens).


%% file: 4_experiments.tex
\section{Experiments}

We evaluate the generated explanations through evaluation of token-level $F_1$ scores comparing them against tokens selected by humans to explain the entailment relation using the e-SNLI dataset \citep{Camburu2018}. The development split of the e-SNLI dataset is used for hyperparameter selection and we report results on the test split. Where multiple annotations are available for a sentence pair, the union of the annotations is taken. We also report average runtime per sentence in seconds measured using 1 thread on an AWS c4.xlarge instance.



\paragraph{Implementation Details}
The model is implemented in AllenNLP \cite{Gardner2018} 
and we optimized our model with Adagrad \cite{Duchi2011}, selecting the models which attained high hypothesis $F_1$ without greatly affecting the accuracy of entailment task (approx 81\% for the thresholded attention model). 
The cell state and hidden dimension was 200 for the LSTM sentence encoder. The projection for attention, $f_{attend}$, was a single layer 200 dimension feed forward network with ReLU activation.
The final feed forward classifier, $f_{cls}$, dimension was $(200,200,3)$ and ReLU activation over the first 2 layers.
For the comparison against black-box explanation mechanisms, we use the code made public by the authors of the respective works setting any hyperparameters to the default values or those suggested in the papers.

\paragraph{Results}
Our experimental results (Table~\ref{tab:res}) indicate that the LIME black-box explanation technique over the model described in Section~\ref{sec:model}
provides token-level explanations that are more similar to human judgments than thresholding the attention distributions. 
We show that the addition of MIL regularizers for generating explanations using thresholded attention improved precision and recall hypothesis explanations. However, similar improvements were not realized for the premise sentence. 
While the black-box methods generated better explanations than thresholded attention, they were 3 orders of magnitude slower.


Only LIME was able to generate good token-level explanations for the premise. This is in contrast to the attention-based explanations of the premise (in the model that LIME was run on) which could not generate satisfactory explanations (see row 2 of Table~\ref{tab:res}). This supports findings in recent works \cite{jain2019attention} that indicate that attention does not always correspond to other measures of feature importance. 
We also found that the black-box model explanation methods behave differently given the same model under test: the premise explanation generated by the Anchors method was more in line with what the model attended to, reflected by the lower recall.


The fully supervised model had high precision yet (relatively) low recall. 
We observed it has a bias towards predicting common words that often appear in highlights (e.g. `man', `woman', `dog', `people') for both premise and hypothesis sentences rather than highlighting keywords that would form an instance-specific explanation.
This behaviour is also more pronounced in the premise sentence highlights rather than the hypothesis. We reason that this due to how the SNLI dataset was constructed: a premise sentence was used to generate 3 hypothesis sentences (entailed, contradicted and neutral). This is corroborated by a survey of 250 instances from the SNLI dataset, where we found that all or part of the subject noun phrase remained unchanged between the premise and hypothesis sentences $60\%$ of the time. 
While the supervised model correctly captured commonly occurring text patterns, as demonstrated by the high $F_1$ scores, this behaviour alone was not sufficient to identify tokens that correlated with the entailment label. We found that most of the commonly predicted tokens by our supervised model did not appear in lists of features highly correlated with the entailment label \cite{Poliak2018,Gururangan2018}. 


%% file: 5_discussion.tex
\section{Conclusions}
In this paper we explored how to generate token-level explanations from NLI models.
We compared the LIME and Anchors black-box methods against a novel, white-box Multiple Instance Learning (MIL) method and a fully supervised baseline.  
The explanations generated by LIME were more similar to the human judgments of the tokens that justify an entailment relation than the attention thresholding approach. This corroborates contemporaneous work \cite{jain2019attention} indicating a lack of correspondence between attention and other measures of feature importance.

The MIL method we introduced steered the attention distributions over tokens in our model to correspond closer to the human judgments allowing better explanations to be generated. Even though, when considering the token-level $F_1$ score, the attention-based explanations were not as good as the black-box techniques we evaluated, they were orders of magnitude faster.


The attention thresholding model we tested did not generate satisfactory explanations had low $F_1$ for the premise sentences.
A possible explanation for the poor performance is what is found by \citet{Rei2018} who show that MIL regularizers performed better when there is a higher degree of association between the sentence-level label and the token-level labels. 
Our model used \emph{independent} encodings of the premise and hypothesis but in NLI there is a strong dependence between the two sentences; thus the entailment prediction should be explained through pairwise token comparisons (e.g.\ synonyms, upward entailment, etc.). 
%
In future work we plan to address this by adding explicit cross-sentence semantic knowledge \cite{joshi2018pair2vec}.